\documentclass[runningheads]{llncs}
\usepackage[T1]{fontenc}
\usepackage{graphicx,verbatim}

\begin{document}
\title{CheXanatomy: Anatomy--Aware Vision--Language Modeling for Chest Radiographs}
\titlerunning{CheXanatomy}

\author{Sergios Gatidis\inst{1,2} \and
Curtis Langlotz\inst{1,2} \and
Christian Bluethgen\inst{1,2}}
\authorrunning{S. Gatidis et al.}
\institute{Stanford Center for Artificial Intelligence in Medicine and Imaging, Stanford University, Palo Alto, CA, USA \and
Department of Radiology, Stanford University, Stanford, CA, USA}

\maketitle              
\begin{abstract}
Vision--language models (VLMs) pretrained on large-scale image–text pairs demonstrate strong image-level understanding, but are primarily optimized for global alignment and do not explicitly encode fine-grained anatomical structure, limiting their suitability for spatially precise tasks such as segmentation.

We introduce CheXanatomy, a framework that integrates explicit anatomical knowledge into a pretrained VLM through autoregressive token-space supervision. Instead of adding task-specific decoder heads, the model is trained to generate anatomical segmentation masks via next-token prediction. To enable scalable supervision, we synthesize realistic chest radiographs from CT volumes and forward-project CT segmentation labels to obtain anatomically consistent 2D masks.

We evaluate the approach on synthetic and real chest radiographs against a U-Net baseline, including ablations on model scale, input resolution, and vision encoder fine-tuning. Autoregressive anatomical supervision achieves performance comparable to specialized convolutional models in-distribution and demonstrates improved geometric robustness under domain shift to real CXR data. In addition, anatomy-pretrained models
exhibit improved sample efficiency when adapting to novel localization
tasks under limited supervision. Larger models and higher input image resolution improve performance, while vision encoder fine-tuning has limited effect.

These results show that embedding anatomical structure directly into the generative objective promotes spatially grounded representations and supports anatomy--aware medical vision–language modeling.


\end{abstract}
%

\section{Introduction}

Accurate anatomical segmentation in chest radiographs (CXR) remains challenging due to projection geometry, overlapping structures, and variability in acquisition protocols. Convolutional models such as U-Net achieve strong in-distribution performance but rely on dense pixel supervision and often degrade under domain shift \cite{ronneberger2015unet,isensee2021nnunet}. In contrast, vision--language models (VLMs) pretrained on image--text pairs learn transferable semantic representations \cite{radford2021clip,zhang2020convirt,tiu2022chexzero}, yet their training objectives emphasize global alignment and do not explicitly encode fine-grained anatomical structure.

We propose to integrate explicit anatomy into VLM training. We introduce CheXanatomy, a framework that trains a pretrained VLM to generate anatomical segmentations autoregressively via next-token prediction, without introducing task-specific pixel-space decoder heads. Segmentation is formulated as structured token generation within the standard generative objective.

Because comprehensive CXR annotations are scarce, we generate scalable supervision by projecting 3D CT segmentations into synthetic radiographs. Using TotalSegmentator and differentiable DRR rendering, we obtain anatomically consistent 2D labels without manual CXR annotation \cite{wasserthal2023totalsegmentator,gopalakrishnan2022fastdrr}.

\begin{figure}[ht]
\centering
\includegraphics[width=\textwidth]{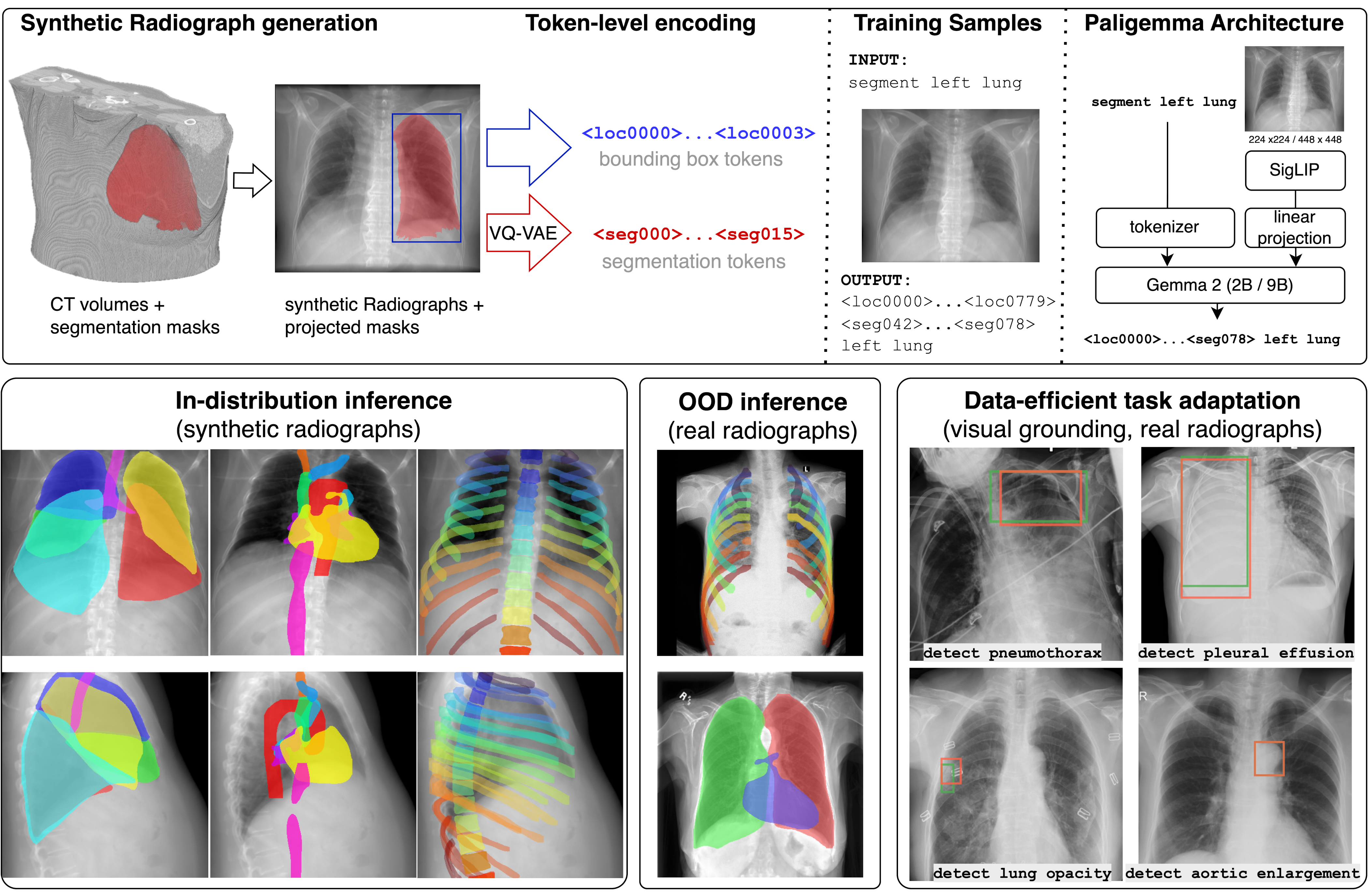}
\caption{Overview of \textbf{CheXanatomy}. CT volumes with anatomic labels are projected into synthetic CXRs, and bounding boxes and segmentation masks are encoded into structured tokens. A pretrained vision--language model (based on the Paligemma architecture) is fine-tuned to autoregressively generate anatomical bounding boxes and segmentations via next-token prediction. The trained model approaches convolutional baselines in-distribution and demonstrates improved geometric robustness under domain shift, while supporting adaptation to new localization tasks. Unlike conventional segmentation networks, no task-specific decoder heads are introduced, and supervision is applied entirely in token space.}
\label{fig:Abstract}
\end{figure}

Our contributions are threefold: 
(1) large-scale synthetic AP and lateral radiographs with projected multi-structure labels; 
(2) autoregressive token-space anatomical supervision of a pretrained VLM via parameter-efficient fine-tuning; 
(3) comprehensive evaluation on synthetic and real radiographs, including ablations on model scale, input resolution, and vision encoder fine-tuning. Unlike projection-based approaches that train standalone segmentation networks, we use synthetic anatomical supervision to reshape the internal representation of a pretrained VLM within its native generative objective (Fig.~\ref{fig:Abstract}).

\section{Related Work}

\paragraph{Anatomical Segmentation on Chest radiographs.}
Early work established benchmarks for lung, heart, and clavicle segmentation in CXR \cite{vanginneken2006chestseg}. Contemporary pipelines rely on U-Net-style architectures and automated configuration strategies \cite{ronneberger2015unet,isensee2021nnunet}. However, large-scale multi-structure annotations remain limited. Recent efforts expand coverage via automated or pseudo-labeling approaches such as CheXmask \cite{gaggion2024chexmask}.

\paragraph{CT-derived Supervision for Radiographs.}
To address annotation scarcity, several works project CT volumes into synthetic radiographs using digitally reconstructed radiographs (DRRs) \cite{hou2024drrrate}. Differentiable DRR frameworks enable principled projection of 3D segmentations to 2D labels \cite{gopalakrishnan2022fastdrr}. CT-projected supervision has been used to obtain detailed anatomical masks and improve robustness in CXR segmentation \cite{seibold2022detailedannotations,seibold2023volumetric,dong2025anycxr}. Our work extends this paradigm by using projection-based anatomy not to train a standalone segmentation model, but to inject structured anatomical knowledge into a pretrained VLM.

\paragraph{Vision--Language Models and Token-Based Dense Prediction.}
Vision--language pretraining (e.g., CLIP) aligns images and text via global objectives \cite{radford2021clip}. Extensions to radiology leverage report supervision but remain largely image-level \cite{zhang2020convirt,tiu2022chexzero}. Promptable segmentation models such as SAM \cite{kirillov2023sam} and CLIP-guided approaches \cite{luddecke2021clipseg,liu2023clipuniversal,koleilat2024medclipsamv2} introduce language-conditioned mask prediction but rely on dedicated decoders.

Recent multimodal foundation models unify visual tasks under autoregressive next-token prediction. Pix2Seq and SegGPT demonstrate dense prediction via sequence modeling \cite{chen2021pix2seq,wang2023seggpt}. PaliGemma supports bounding box and segmentation outputs through structured token generation within a unified VLM framework \cite{beyer2024paligemma}. However, explicit anatomical supervision for radiography within such autoregressive VLMs remains underexplored.

Our work leverages PaliGemma's native token interface to encode fine-grained anatomy directly into the generative objective using scalable CT-derived supervision.

\section{Methods}

\paragraph{Autoregressive Anatomical Segmentation.}
We used PaliGemma 2 \cite{steiner2024paligemma2}, a pretrained vision--language model that supports structured spatial outputs, including bounding boxes and segmentation masks, through autoregressive next-token prediction. PaliGemma 2 consists of a SigLIP-So400m vision encoder \cite{zhai2023siglip} with a linear projection into the language token space and a Gemma 2 \cite{gemmateam2024gemma2} large language model for autoregressive prediction. Bounding boxes are represented using four dedicated tokens. Segmentation masks are represented using 16 mask tokens per segment.
Mask prediction is implemented through a vector-quantized variational autoencoder (VQ-VAE) trained to encode binary masks into a discrete 16-dimensional latent representation. During training, the model predicts these discrete latent tokens, which are then decoded into segmentation masks using the VQ-VAE decoder. Optimization is performed entirely in token space using standard next-token cross-entropy loss. No pixel-level loss, auxiliary decoder, or additional segmentation objective is introduced. This formulation enables anatomical supervision within the generative training paradigm of the pretrained model.

\paragraph{Synthetic CXR Generation from CT.}
Large-scale anatomical annotations for chest radiographs are limited. To address this, we generated synthetic projection radiographs from 3D CT volumes. We used non-contrast CT scans from the CT-RATE dataset \cite{hamamci2025ctrate}, consisting of 12,984 training CT volumes and 683 randomly selected test CT volumes.
For each CT volume, two digitally reconstructed radiographs were generated: one posterior--anterior projection and one lateral projection. Forward projection was performed using Diff-DRR \cite{gopalakrishnan2022fastdrr}. This resulted in 25,968 synthetic training images and 1,366 synthetic test images.
Multi-structure anatomical segmentations were obtained from each CT volume using the TotalSegmentator \cite{wasserthal2023totalsegmentator}. The 3D segmentation labels were projected forward into the 2D projection domain to obtain anatomically consistent segmentation masks aligned with the synthetic radiographs. The binary masks were encoded into the VQ-VAE latent token space required for autoregressive prediction.
Training supervision covered 80 anatomical targets spanning thoracic skeletal anatomy, pulmonary structures, cardiac chambers and major vessels, and selected upper abdominal organs visible in projection. Multiple textual synonyms were sampled during training to construct prompts for each anatomical label.
To increase robustness, random scaling transformations were applied to synthetic training images to simulate variability in anatomy size and acquisition geometry. This projection-based strategy scales anatomical supervision without requiring manual CXR annotation.

\paragraph{Parameter-Efficient Fine-Tuning.}
Model adaptation was performed using Low-Rank Adaptation (LoRA) \cite{hu2021lora}. This enabled parameter-efficient fine-tuning while keeping most pretrained weights fixed. Two configurations were evaluated: fine-tuning with the vision encoder frozen, and fine-tuning with the vision encoder updated. Experiments were conducted using 3B and 10B parameter models and input resolutions of $224\times224$ and $448\times448$. Training was performed on 8 H100 GPUs with batch sizes between 8 (10B parameter model with $448\times448$ image resolution) and 24 (3B parameter model with $224\times224$ image resolution).

\paragraph{Convolutional Baseline.}
As a supervised baseline, we trained a two-dimensional U-Net using the same synthetic training data. The network consisted of four encoder stages. Each stage included two $3\times3$ convolutional layers followed by batch normalization and ReLU activation, and a $2\times2$ max pooling layer for downsampling. The number of feature channels doubled at each stage starting from 32, resulting in feature dimensions of 32, 64, 128, and 256, with a bottleneck of 512 channels. The decoder used $2\times2$ transposed convolutions for upsampling and concatenated skip connections from corresponding encoder features. A final $1\times1$ convolution produced one output channel per anatomical class (80 in total). The input resolution was $448\times448$.

\paragraph{Evaluation Data and Metrics.}
Segmentation performance was evaluated on three datasets: synthetic CT-RATE test projections (1,366 images from posterior--anterior and lateral views), CheXmask \cite{gaggion2024chexmask} (200 randomly selected posterior--anterior radiographs with heart and lung masks), and VinDr-RibCXR \cite{nguyen2021vindr_ribcxr} (198 posterior–anterior radiographs with bilateral rib segmentations). 
Segmentation quality was assessed using Dice coefficient, Intersection-over-Union (IoU), Hausdorff distance, and Fourier descriptor distance (FDD). Bounding box localization was evaluated using bounding box IoU.

\paragraph{Ablation Studies.}

Ablation experiments were performed to assess the impact of model scale, input resolution, and vision encoder fine-tuning. Comparisons were made between 3B and 10B parameter models, between $224\times224$ and $448\times448$ input resolutions, and between frozen and fine-tuned vision encoders.

\paragraph{Data-Efficient Transfer to Novel Localization Tasks.}

To evaluate transfer to novel localization tasks, few-shot experiments were conducted on visual grounding tasks from the RadVLM dataset \cite{deperrois2025radvlm}. Baseline models as well as anatomy-pre-trained models were fine-tuned for one epoch using subsets of labeled data corresponding to fractions between  0\% and 100\% of the available RadVLM training data (n=118,360). Performance on held-out test data (n=22,972) was measured using bounding box Intersection-over-Union and compared to models without anatomical pretraining.

\paragraph{Code Availability.}

Code is available at:

\begin{itemize}

  \item \url{https://github.com/sergiosgatidis/CheXanatomy}

  \item \url{https://github.com/sergiosgatidis/CheXsynth}

\end{itemize}

\begin{figure}[ht]
\centering
\includegraphics[width=\textwidth]{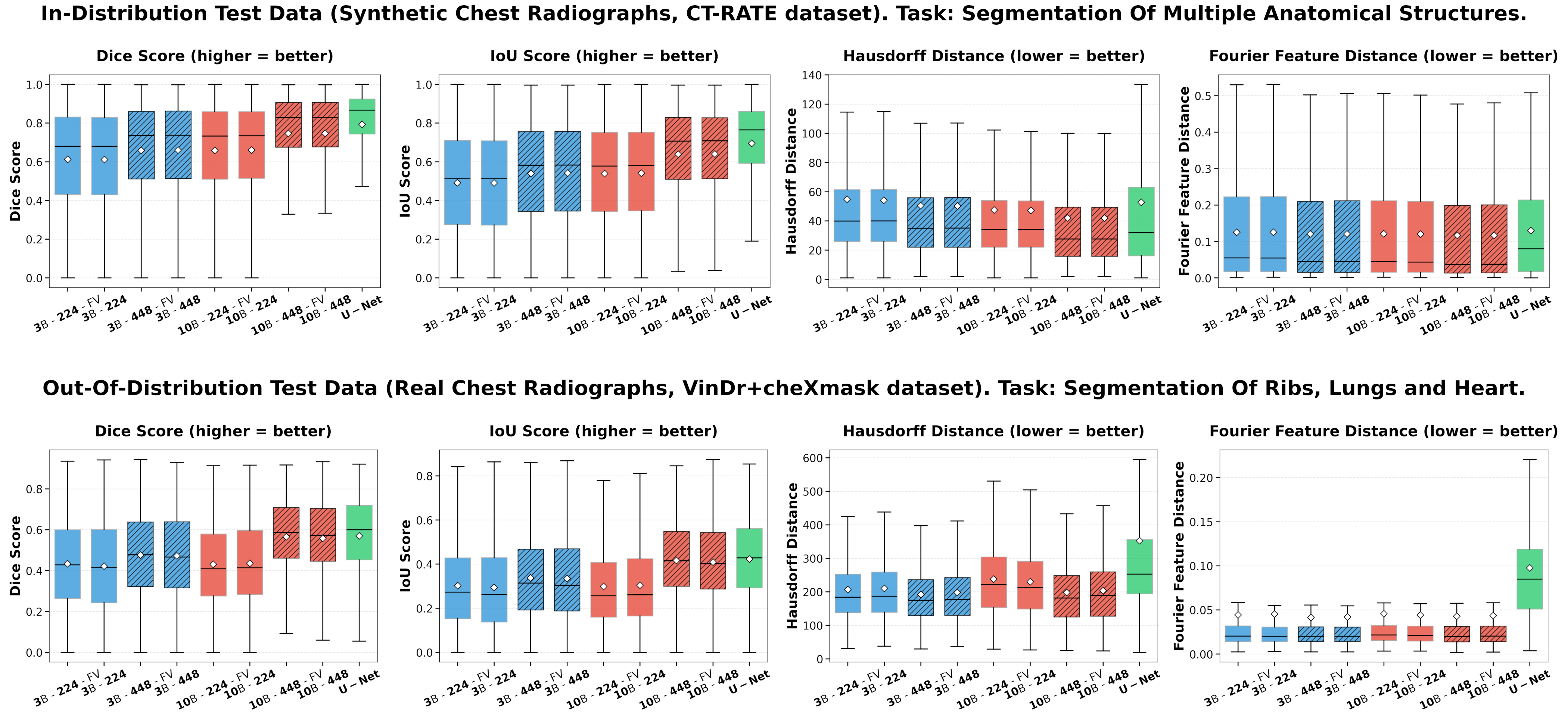}
\caption{Segmentation performance on synthetic CT-RATE (top) and combined real-world radiographs from CheXmask and VinDr-RibCXR (bottom). Boxplots show Dice, IoU, Hausdorff distance, and Fourier descriptor distance across model scales and input resolutions. On synthetic data, the best PaliGemma models approach U-Net in overlap metrics and achieve lower boundary and shape errors. Under domain shift, PaliGemma maintains comparable Dice and IoU while consistently reducing boundary errors relative to U-Net. Models are named based on size (3B vs. 10B) and input image size ($224\times224$ vs. $448\times448$). FV = Vision encoder frozen during training.}
\label{fig:Test}
\end{figure}

\section{Results}

\paragraph{In-Distribution Performance on synthetic CXRs.}
Segmentation performance on the held-out synthetic CT-RATE radiographs is summarized in Fig.~\ref{fig:Test} (top). The best PaliGemma configuration (10B, $448\times448$) approached the performance of the specialized U-Net baseline in overlap-based metrics, achieving comparable Dice and IoU values.
While U-Net achieved slightly higher Dice and IoU overall, the high-resolution 10B PaliGemma model obtained lower Hausdorff distance and lower Fourier descriptor distance, indicating improved boundary alignment and shape consistency. Increasing input resolution consistently improved segmentation performance across both model scales. Fine-tuning the vision encoder had no measurable influence on performance.

\begin{figure}[ht]
\centering
\includegraphics[width=\textwidth]{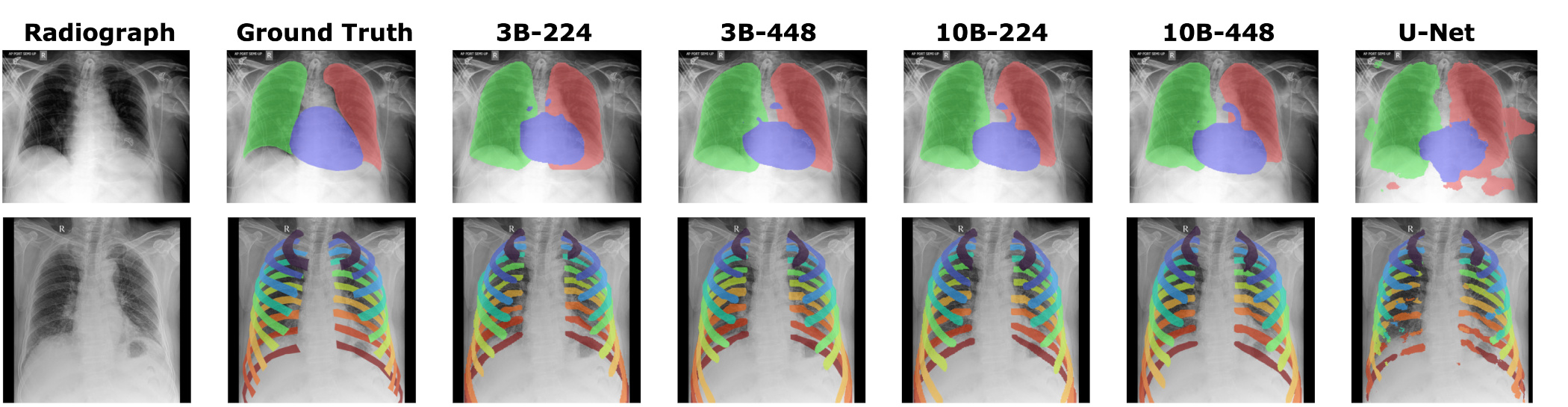}
\caption{Representative examples for out-of-distribution segmentation performance of anatomy-trained VLMs and a specialized U-Net compared to ground truth segmentations. Only VLMs with frozen vision encoders during fine-tuning are shown. Top row: CheXmask example (lungs and heart). Bottom row: VinDr-RibCXR example (bilateral rib structures). Compared to U-Net, anatomy-trained VLMs demonstrate more stable geometric structure and reduced boundary irregularities under domain shift.}
\label{fig:OOD}
\end{figure}

\paragraph{Out-of-Distribution Generalization on Real CXRs.}
Combined results across CheXmask and VinDr-RibCXR are shown in Figs. ~\ref{fig:Test} (bottom) and ~\ref{fig:OOD}. A consistent and practically relevant pattern emerges under domain shift. The best PaliGemma models achieved Dice and IoU comparable to the U-Net baseline while consistently outperforming it on boundary and shape-based metrics.
In particular, Hausdorff distance and Fourier descriptor distance were substantially lower for PaliGemma models compared to U-Net, indicating improved geometric stability and shape fidelity on real radiographs. The 10B model at $448\times448$ resolution closely matched U-Net in overlap metrics while maintaining superior boundary and shape consistency.

\paragraph{Data-Efficient Transfer to Novel Localization Tasks.}

\begin{figure}[ht]
\centering
\includegraphics[width=\textwidth]{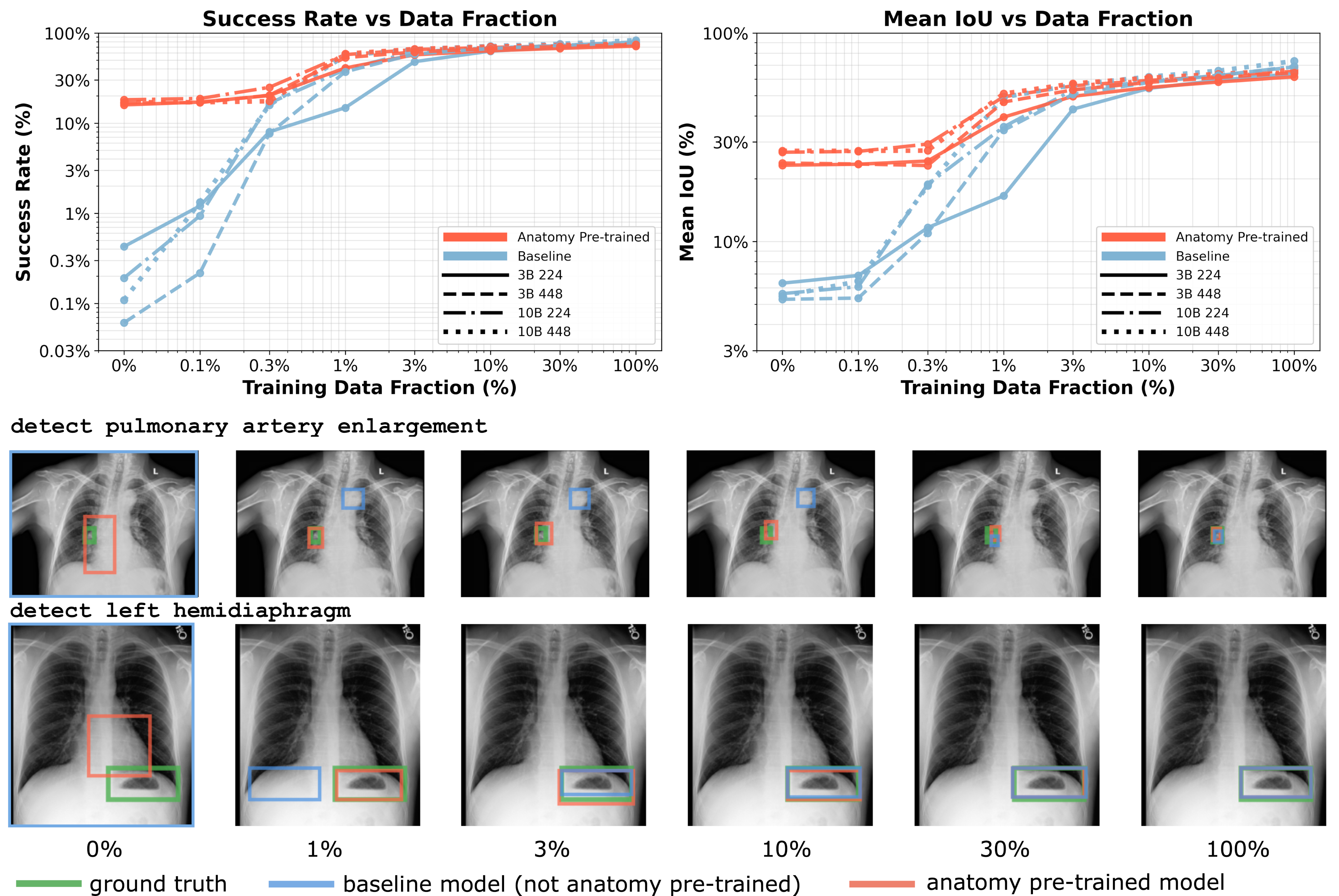}
\caption{Performance scaling on the RadVLM grounding benchmark comparing anatomy-pretrained and baseline models across training data fractions. Top: Success rate and mean IoU as a function of available training data (0--100\%). Anatomy-pretrained models show clear advantages in the low-data regime (0--3\%), with performance converging as more task-specific data becomes available. Bottom: Representative grounding examples for two tasks (``detect pulmonary artery enlargement'' and ``detect left hemidiaphragm'') across data fractions. Green: ground truth, red: anatomy-pretrained model (3B-448), blue: baseline model (3B-448). Anatomy pretraining provides improved localization accuracy and stability when supervision is limited.}
\label{fig:RadVLM}
\end{figure}

We evaluated whether ana-tomy pretraining improves adaptation to unseen localization tasks using the RadVLM visual grounding dataset \cite{deperrois2025radvlm}. When fine-tuned with limited or no supervision (0\%, 1\%, and 3\% of the training data), anatomy-pretrained models consistently outperformed their non-pretrained counterparts in bounding box IoU, as shown in Fig.~\ref{fig:RadVLM}. The performance gap was most pronounced in the low-data regime, indicating improved sample efficiency and a stronger spatial prior. As the amount of task-specific training data increased (10\% and above), performance between pretrained and non-pretrained models converged. These results suggest that explicit anatomical supervision provides a structured initialization that facilitates rapid adaptation to new spatial tasks, particularly when labeled data are scarce.

\section{Discussion}

We demonstrated that explicit anatomical supervision can be integrated into a pretrained vision--language model through autoregressive token prediction. Our results show that token-space supervision enables segmentation performance comparable to a specialized convolutional baseline on synthetic in-distribution data, while improving geometric robustness under domain shift to real radiographs.

On synthetic CT-RATE projections, the best PaliGemma configurations approached U-Net in Dice and IoU and achieved lower boundary and shape errors. Under real-world distribution shift, PaliGemma maintained comparable overlap metrics while consistently reducing Hausdorff and Fourier descriptor distances. These findings indicate that anatomical supervision in token space promotes spatially coherent and geometrically stable representations, even when trained solely on synthetic projections.

Scaling model size and input resolution improved performance, whereas fine-tuning the vision encoder had limited impact, suggesting that anatomical reasoning primarily emerges in the multimodal token space. Few-shot localization experiments further demonstrated improved sample efficiency, with anatomy-pretrained models outperforming baselines in low-data regimes and converging as task-specific supervision increased.

Beyond quantitative performance, the autoregressive VLM formulation offers structural advantages over conventional segmentation networks. The same model supports flexible language prompting, enables integration of new anatomical structures through lightweight fine-tuning without architectural modification, and can be embedded into pipelines requiring spatial grounding, such as grounded report generation or region-level reasoning.

Overall, structured anatomical supervision provides a scalable mechanism for aligning generative medical vision--language models with spatial structure in radiographs.

\section{Conclusion}

We introduced CheXanatomy, a framework for injecting explicit anatomical knowledge into a pretrained vision--language model through autoregressive token-space supervision. By leveraging scalable CT-derived synthetic radiographs, we trained a VLM to generate anatomical segmentations without task-specific decoder heads. 

Our results demonstrate that anatomy-aware token supervision enables competitive in-distribution performance and improved geometric generalization to real radiographs. More broadly, this suggests that spatial grounding can be embedded directly into multimodal foundation models through structured generative supervision.
\clearpage

\bibliographystyle{splncs04}

\bibliography{references}

\end{document}